\documentclass[11pt]{article}

\usepackage[utf8]{inputenc}
\usepackage{amsmath}
\usepackage{amsfonts}
\usepackage{amssymb}
\usepackage{amsthm}
\usepackage{graphicx}
\usepackage{caption}
\usepackage{xcolor}
\usepackage{float}
\usepackage[hidelinks,colorlinks=true,linkcolor=blue,citecolor=blue]{hyperref}
\usepackage{natbib}
\usepackage{verbatim}
\usepackage{mathtools}
\usepackage{tikz}
\usepackage{bbm}
\usepackage{subcaption}
\usepackage{authblk}
\theoremstyle{plain}

\newcommand{\reals}{\mathbb{R}}
\newcommand{\naturals}{\mathbb{N}}

\newcommand{\Fcal}{\mathcal{F}}

\newcommand{\Xcal}{\mathcal{X}}

\newcommand{\Vcal}{\mathcal{V}}
\newcommand{\Wcal}{\mathcal{W}}
\newcommand{\Pcal}{\mathcal{P}}

\newcommand{\Ecal}{\mathcal{E}}

\DeclareMathOperator*{\argmax}{arg\,max}
\DeclareMathOperator*{\argmin}{arg\,min}
\DeclareMathOperator*{\expect}{\mathbb{E}}

\newcommand{\identity}{\mathbf{I}}
\newcommand{\imaginary}{\operatorname{i}}

\newcommand{\norm}[1]{\left\lVert#1\right\rVert}
\newcommand{\inner}[2]{\left\langle #1, #2 \right\rangle}

\usepackage[letterpaper,top=2cm,bottom=2cm,left=3cm,right=3cm,marginparwidth=1.75cm]{geometry}

\title{Operator Learning: A Statistical Perspective}

\author[ ]{Unique Subedi, Ambuj Tewari}
\affil[ ]{Department of Statistics, University of Michigan}
\affil[ ]{\texttt{\{subedi, tewaria\}@umich.edu}}
\date{}

\begin{document}

\maketitle
\begin{abstract}
Operator learning has emerged as a powerful tool in scientific computing for approximating mappings between infinite-dimensional function spaces. A primary application of operator learning is the development of surrogate models for the solution operators of partial differential equations (PDEs). These methods can also be used to develop black-box simulators to model system behavior from experimental data, even without a known mathematical model. In this article, we begin by formalizing operator learning as a function-to-function regression problem and review some recent developments in the field. We also discuss PDE-specific operator learning, outlining strategies for incorporating physical and mathematical constraints into architecture design and training processes. Finally, we end by highlighting key future directions such as active data collection and the development of rigorous uncertainty quantification frameworks.
\end{abstract}

\section{INTRODUCTION}

Artificial Intelligence (AI) is making rapid advances in a variety of areas ranging from language and vision modeling to applications in the physical sciences. AI is transforming scientific research and accelerating scientific discovery by providing new tools for modeling complex systems \citep{tang2020introduction} and optimizing workflows \citep{wang2023scientific}. A notable example is the protein structure prediction with AlphaFold \citep{jumper2021highly}, which was awarded the 2024 Nobel Prize in Chemistry \citep{nobel2024chemistry}. This growing impact has led to the emergence of ``AI for Science" \citep{zhang2023artificial}, a paradigm that seeks to integrate AI into scientific problem-solving. The subject of this article, operator learning, is one of the key approaches within this new paradigm.

In mathematics, a mapping between infinite dimensional function spaces is often called an {\em operator}. Operator learning is an area at the intersection of applied mathematics, computer science, and statistics which studies how we can {\em learn} such operators from data. Its primary application is the development of fast and accurate surrogate models \citep{bhattacharya2021model} for the solution operators of partial differential equations (PDEs). Additionally, as a data-driven approach, operator learning techniques can be used to develop black-box simulators that simulate system behavior based on observed experimental data \citep{you2022physics, you2022learning}, even when the underlying mathematical model is unknown. 

Before formally defining the problem of operator learning, let us discuss a motivating example that illustrates its relevance. Many physical systems are governed by PDEs, which describe how the system evolves given particular initial conditions. A classic example is the heat equation 
\citep[Section 2.3]{evans2022partial}
\begin{equation}\label{eq:heat} \frac{\partial u}{\partial t} = \tau \nabla^2 u, \end{equation} 
that arises in heat conduction and diffusion problems.
Here, $\tau > 0$ could be the thermal conductivity of a material,  $u: \Xcal \times [0,\infty) \to \reals$ for some set $\Xcal \subseteq \reals^d$ could define a temperature profile at any given space-time coordinate, and $\nabla^2$ is the Laplacian operator defined as $\nabla^2 u := \sum_{j=1}^d \partial^2 u/ \partial x_j^2$. The solution of the heat equation can be written using a linear operator defined as
\[ \exp(\tau t \nabla^2) : = \sum_{k=0}^\infty \frac{(\tau t \,\nabla^2)^k}{k!}.\]
That is, given an initial condition $u_0$, the solution function can be written as $u_t = \exp(\tau t \nabla^2) u_0$ for any time point $t>0$ \citep[Chapter 5.4]{hunter2023pde}.

This solution operator is useful primarily for conceptual understanding and cannot be used to obtain the solution function in all but a few cases of simple domain geometry and simple initial conditions. The solution is generally obtained using PDE solvers which use numerical methods to map the initial conditions $u_0$ to $u_t$ at some desired time point $t>0$.
Such solver starts from scratch for every new initial condition $u_0$ of interest. Since the solver is computationally slow and expensive, this ab initio approach to evaluating solutions can be limiting in applications such as engineering design where the solution needs to be evaluated for many different initial conditions \citep{umetani2018learning}. To solve this problem, operator learning aims to learn the solution operator directly from the data. By amortizing the computational cost through upfront training, these learned operators allow for significantly efficient solution evaluation compared to traditional solvers while sacrificing a small degree of accuracy.

\begin{figure}[h]
    \centering
    \begin{subfigure}[b]{0.32\textwidth}
        \centering
        \includegraphics[width=\textwidth]{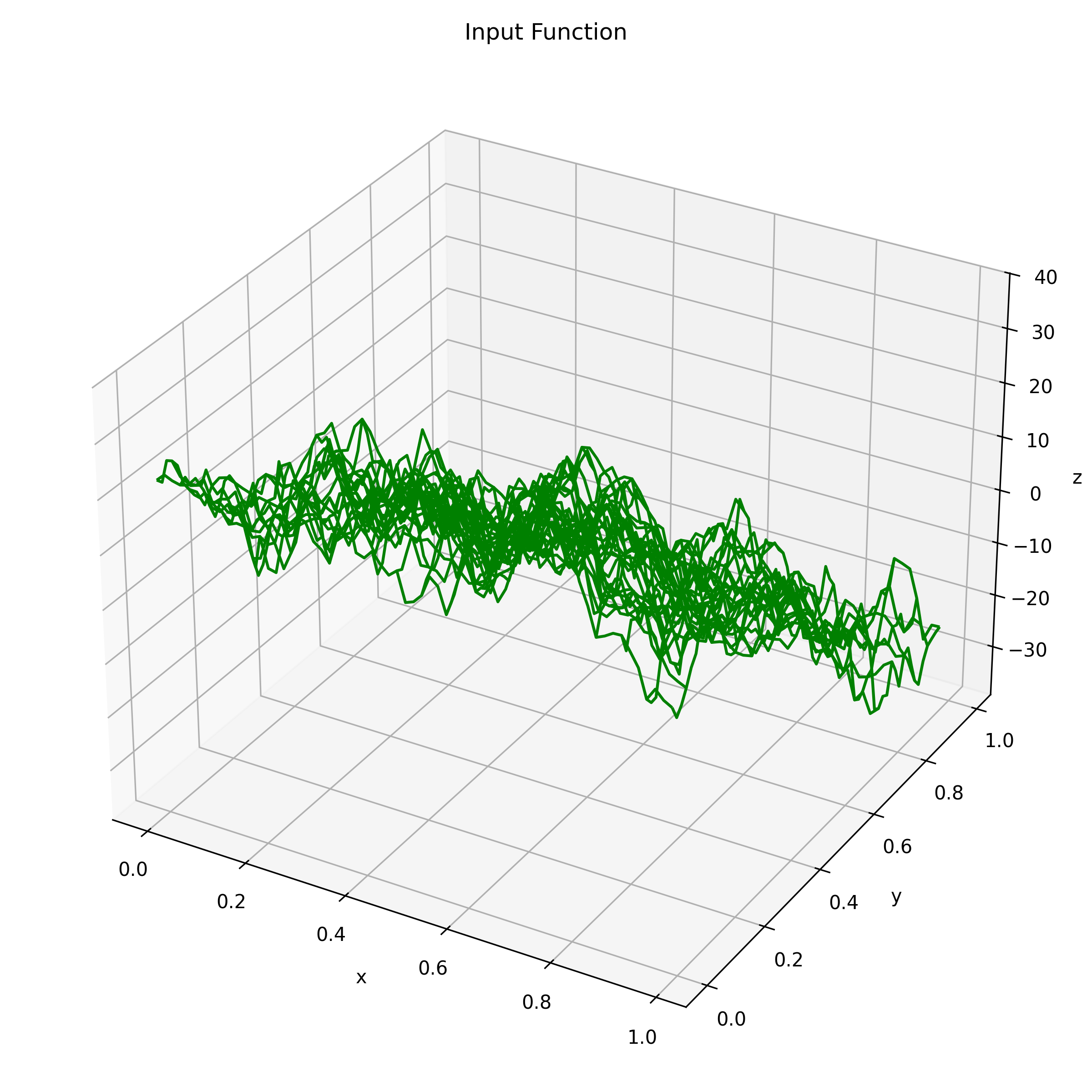}
        \caption{Input ($v$)}
    \end{subfigure}
    \begin{subfigure}[b]{0.32\textwidth}
        \centering
        \includegraphics[width=\textwidth]{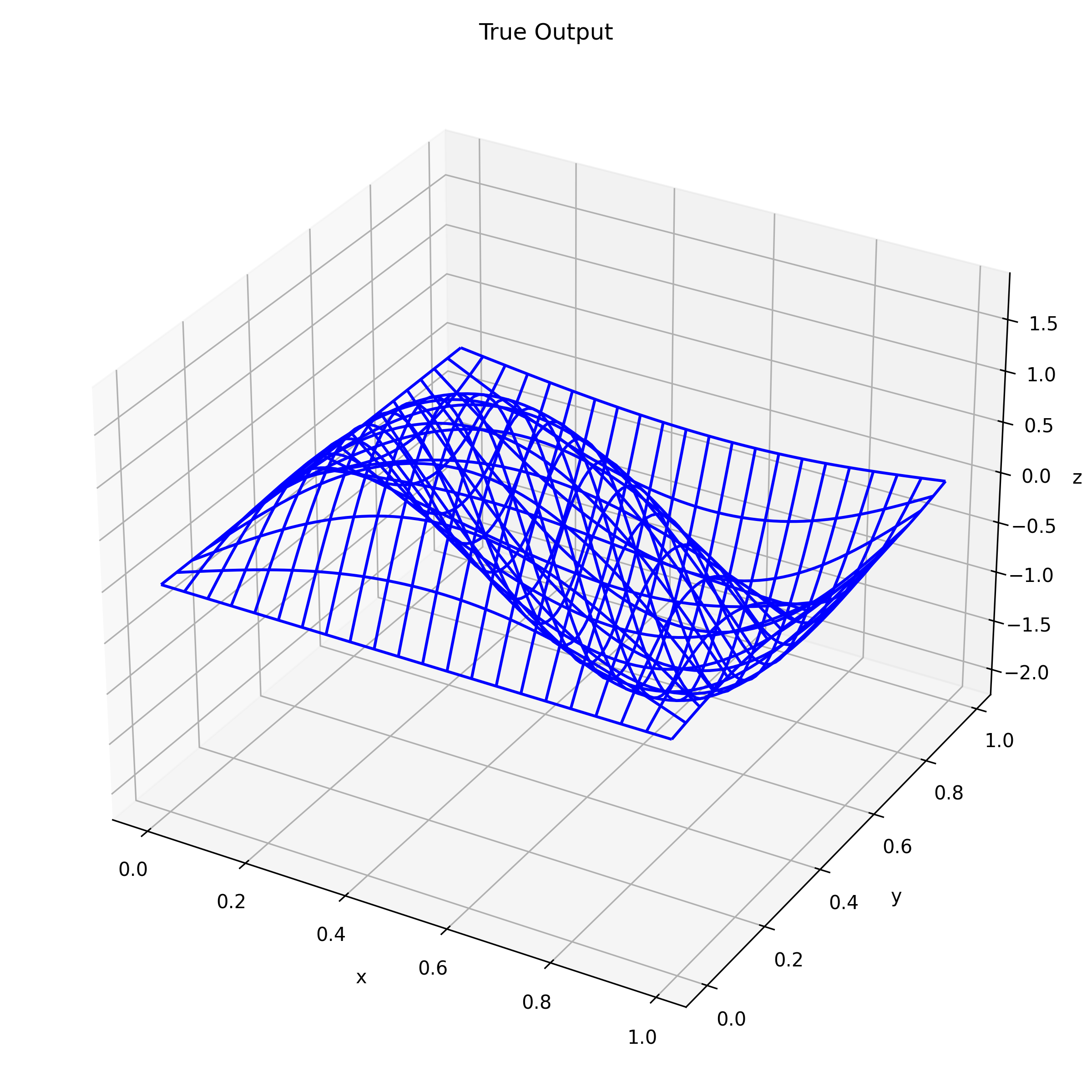}
        \caption{Output ($w$) }
    \end{subfigure}
    \begin{subfigure}[b]{0.32\textwidth}
        \centering
        \includegraphics[width=\textwidth]{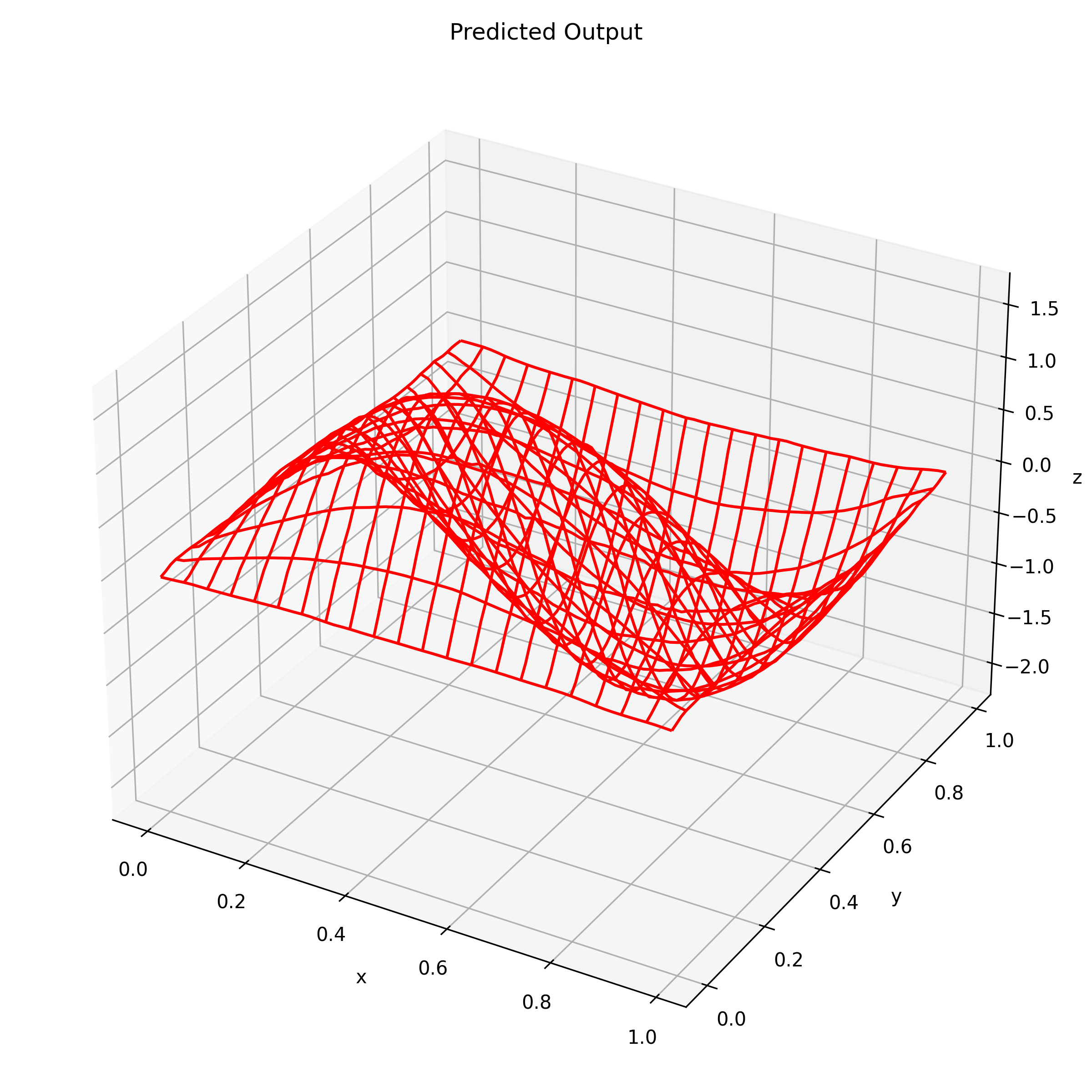}
        \caption{Prediction ($\widehat{w}$)}
    \end{subfigure}
    \vspace{0.2cm}
    \caption{Input function, ground truth solution of the heat equation, and the predicted solution by a Fourier Neural Operator. Here, we use $ \Xcal = [0,1)^2, \tau=0.05$, and $T=1$. The corresponding code is available in this \href{https://github.com/unique-subedi/ARSIA-review/blob/main/ARSIA_review_OL_statistical.ipynb}{notebook}. } 
    \label{fig:heat}
\end{figure}

More precisely, for some prespecified time point $t=T$, let $ \operatorname{G} :=  \exp(\tau T \nabla^2)$ denote the solution operator of interest. Then, given training data $(v_1, w_1), \ldots, (v_n, w_n)$ where $v_i$ is the initial condition and $w_i = \operatorname{G}(v_i) $ is the solution at time point $T$, operator learning involves estimating an approximation $ \widehat{\operatorname{F}}_n $ of $ \operatorname{G} $ by searching over a  a predefined operator class $ \mathcal{F} $ \citep[Section 2]{kovachki2024operator}. Once trained, the estimated operator can be used to predict an approximate solution $ \widehat{w} = \widehat{\operatorname{F}}_n(v) $ for a new initial condition $ v $. The objective is to design an estimation procedure such that $ \widehat{w} $ closely approximates the true solution $ w = \operatorname{G}(v) $ under a suitable metric. For illustration, Figure \ref{fig:heat} compares the Fourier neural operator's prediction and the actual output from a PDE solver.

In the example of the heat equation, we wanted to learn an operator that maps an initial condition to the solution function. But operator learning methods can also be used to learn how the parameters of steady-state equations map to their corresponding solution functions. For example, for a prototypical elliptic PDE
\[
-\text{div}(v(\cdot) \, \nabla w) = f, \quad \quad v(x)=0  \quad \forall x \in \text{boundary}(\Xcal),
\]
 one may seek to learn the nonlinear ground truth operator $\operatorname{G}$ that maps the coefficient function $ v $ to the solution function $ w $. The function $ v:\Xcal \to [0, \infty] $ typically represents permeability of the medium in fluid flow applications \citep{cheng1984darcy}. A surrogate operator trained on for this mapping can efficiently predict how the solution function $ w $ changes in response to variations in the system’s properties defined by $v$.

Additionally, these operator surrogates can also be used to complement traditional PDE solvers rather than completely replace them. For example, the prediction of operator surrogate can be used as an initialization for spectral solvers of nonlinear PDEs, which often rely on iterative methods such as Newton’s method \citep{AGHILI2025108434}. A well-chosen initialization can substantially accelerate convergence, reducing the overall computational cost.

Although operator learning is a relatively new field, it has already had practical impact. For example, the European Centre for Medium-Range Weather Forecasts has incorporated an operator learning model, FourcastNet, into its experimental forecasting suite \citep{ecmwf_fourcast_2025}. In the medical field, operator learning has been applied to optimize catheter prototypes—devices used in surgical procedures and disease treatment \citep{zhou2024ai}. Similarly, it has been used to model plasma evolution in nuclear fusion research \citep{gopakumar2024plasma}.

Given these early successes, operator learning not only received coverage in public media \citep{ananthaswamy2021, anandkumar2023ai} but also attracted the attention of several research communities. For example, \cite{kovachki2024operator} explore the foundations operator learning from an approximation theory perspective, \cite{boulle2023mathematical} discuss its connections to numerical linear algebra, and \cite{azizzadenesheli2024neural} provide an application-focused overview. Our article contributes to this growing body of work by adopting a statistical perspective on operator learning.

We begin by formalizing operator learning as a regression problem between function spaces, a problem that has been studied by the statistics community as ``functional regression" or ``function-on-function regression" within the functional data analysis (FDA) literature. Recent advances in data-driven surrogate modeling for PDEs have revitalized the interest in this problem. We adopt the term ``operator learning" instead of ``functional regression" as it better reflects our viewpoint of using statistical learning framework with an appropriate loss function and a class of operators. Next, we review recent progress in operator learning, and finally, we conclude by discussing key open questions and future research directions.

\section{A STATISTICAL FRAMEWORK FOR OPERATOR LEARNING}

We approach operator learning from the point of view of statistical learning theory~\citep{vapnik2000nature} with the important difference that both input and output spaces are spaces consisting of {\em functions}. To this end, fix $\mathcal{X} \subseteq \mathbb{R}^d$, and let $\Vcal, \Wcal$ to be separable Banach spaces of $\reals^p$-valued functions on $\Xcal$. As an example, one may have 
 $ \Vcal = \Wcal = L^2_{\nu}(\mathcal{X}, \mathbb{R}^p)$, the set of square-integrable $\mathbb{R}^p$-valued functions defined on the domain $\mathcal{X}$. The base measure $\nu$ is typically the Lebesgue measure; however, for certain cases, such as when $\mathcal{X} = \mathbb{R}^d$, a weighted measure like Gaussian (with density proportional to $e^{-\alpha^2 \|x\|^2} \, dx$) may be considered to ensure the base measure is finite. In general, $\mathcal{V}$ can be defined as functions mapping from $\mathbb{R}^{d_1}$ to $\mathbb{R}^{p_1}$, and $\mathcal{W}$ as functions mapping from $\mathbb{R}^{d_2}$ to $\mathbb{R}^{p_2}$. However, the learning-theoretic and practical aspects of the problem remain largely unchanged under such generality.  Thus, we take $d_1=d_2$ and $p_1=p_2$ to minimize notational complexity and simplify exposition.

We now formally define the statistical problem of operator learning. Let $\operatorname{G}:\Vcal \to \Wcal$ be the ground truth operator of interest. For example, this could be the solution operator of the PDE of interest. In the statistical learning framework, the learner is provided with $n$ i.i.d. samples $\{(v_i,\operatorname{G}(v_i))\}_{i=1}^n$, where $v_i$'s are drawn iid from a distribution $\mu$. Throughout the work, we will assume that $\mu$ belongs to a family of distributions $\Pcal$ on $\Vcal$, and the learner has knowledge of the family.

Using this sample and a predefined learning rule, the learner constructs an estimator $\widehat{\operatorname{F}}_n$. For simplicity, we use $\widehat{\operatorname{F}}_n$ to denote both the estimator and the estimation rule. Although $\widehat{\operatorname{F}}_n$ can theoretically be any operator from $\mathcal{V}$ to $\mathcal{W}$, in practice, the learner typically searches within a prespecified operator class $\mathcal{F} \subseteq \mathcal{W}^\mathcal{V}$. For example, $\mathcal{F}$ might be the class of bounded linear operators or neural network-based operator classes (see Section \ref{sec:op-classes}). However, the true operator of interest $\operatorname{G}$ need not be in the class $\Fcal$. Accordingly, the estimator $\widehat{\operatorname{F}}_n$ is evaluated relative to the best operator within the class $\mathcal{F}$. Thus, for a prespecified loss function $\ell: \mathcal{W} \times \mathcal{W} \to \mathbb{R}_{\geq 0}$, the worst-case (over $\mu \in \Pcal$) expected  excess risk of $\widehat{\operatorname{F}}_n$ is defined as
\begin{equation}\label{eq:excess-risk}
    \mathcal{E}_n(\widehat{\operatorname{F}}_n, \mathcal{F}, \Pcal, \operatorname{G}) := \sup_{\mu \in \Pcal} \left(\expect_{v_1, \ldots, v_n \sim_{\text{iid}} \, \, \mu} \left[ \expect_{v \sim \mu} \big[ \ell\big(\widehat{\operatorname{F}}_n(v), \operatorname{G}(v)\big) \big] \right] - \inf_{\operatorname{F} \in \mathcal{F}} \expect_{v \sim \mu} \big[ \ell\big(\operatorname{F}(v), \operatorname{G}(v) \big) \big] \right).
\end{equation}
Then, the learner’s objective is to develop an estimation rule such that $  \mathcal{E}_n(\widehat{\operatorname{F}}_n, \mathcal{F}, \Pcal, \operatorname{G}) \to 0$ as $n \to \infty$. 

To make the problem non-trivial from an estimation perspective, we assume that the learner does not know the exact form of $ \operatorname{G} $. Otherwise, the learner could simply set $ \widehat{\operatorname{F}}_n = \operatorname{G} $ or its closest approximation in $ \mathcal{F} $, making the estimation problem trivial. However, the learner may still have some structural knowledge about $ \operatorname{G} $, such as its linearity or smoothness properties, based on prior information of the underlying PDE. Alternatively, the learner may know that the class $ \mathcal{F} $ contains an operator that approximates $ \operatorname{G} $ to within a small error, meaning that  
\[
\inf_{\operatorname{F} \in \mathcal{F}} \expect_{v \sim \mu} \big[ \ell\big(\operatorname{F}(v), \operatorname{G}(v) \big) \big] \leq \varepsilon
\]
for some small $ \varepsilon > 0 $. This assumption is reasonable for function classes with a universal approximation property.  In statistical learning, this is often referred to as the well-specified or $ \varepsilon $-realizable setting. 
The well-known additive noise model is a special case of this setting. To see this, let \(\operatorname{G}(v) = \operatorname{F}^{\star}(v) + \delta\) for some \(\operatorname{F}^{\star} \in \mathcal{F}\) and a random noise term \(\delta \in \mathcal{W}\) such that $\expect[\norm{\delta}_{\Wcal}^2]\leq \varepsilon$. Here, instead of $\operatorname{G}$ being the ground truth, it is a $\delta$ perturbation of the output of ground truth $\operatorname{F}^{\star}$.
If the loss function \(\ell\) is the squared norm in the Banach space \(\mathcal{W}\), then $
\inf_{\operatorname{F} \in \mathcal{F}} \mathbb{E} \left[\|\operatorname{F}(v)- \operatorname{G}(v) \|_{\mathcal{W}}^{2} \right] \leq \mathbb{E} \left[\|\operatorname{F}^{\star}(v)- \operatorname{G}(v) \|_{\mathcal{W}}^{2} \right] = \mathbb{E}[\|\delta\|_{\mathcal{W}}^2] \leq \varepsilon.
$ Here, the expectation accounts for the randomness in both \( v \) and \( \delta \).


\subsection{Loss Functions}\label{sec:loss}
A common choice for the loss function is $ \ell(\widehat{w}, w) = \|\widehat{w} - w\|_{\mathcal{W}}^q $, where $ \|\cdot\|_{\mathcal{W}} $ denotes the canonical norm associated with the Banach space $ \mathcal{W} $. In practice, $ q $ is typically set to 1 or 2. Another frequently used loss function is the relative loss, given by  
\[
\ell(\widehat{w}, w) = \frac{\|\widehat{w} - w\|_{\mathcal{W}}^q}{\|w\|_{\mathcal{W}}^q},
\]
which is often preferred in empirical settings, as it has been observed to yield better results \citep[Section 6.5]{kovachki2023neural}. Additionally, if prior knowledge about the smoothness of the functions is available, $ \mathcal{W} $ may be chosen as an appropriate Sobolev space to incorporate this information into the learning process, generally referred to as Sobolev training \citep{son2021sobolev}.

\subsection{Distribution Families}
In principle, one could define $ \mathcal{P} $ as the set of all Borel probability distributions on $ \mathcal{V} $. This family with a worst-case analysis over all operators $ \operatorname{G}: \mathcal{V} \to \mathcal{W} $ leads to the standard minimax framework. However, minimax analysis fails to capture the prior knowledge that a practitioner generally has. Thus, it is common to impose additional constraints on $ \mathcal{P} $.

In the applied operator learning literature, $ \mathcal{P} $ is often chosen as a class of Gaussian process with specific covariance structures. Works such as \cite{bhattacharya2021model, li2020fourier, kovachki2023neural} assume that input functions are drawn from a Gaussian process $ \text{GP}(0, \alpha(-\nabla^2 + \beta \mathbf{I})^{-\gamma}) $ for some $ \alpha, \beta, \gamma > 0 $. This assumption is justified in these works as the datasets are synthetically generated by directly sampling functions from this distribution. Similarly, \cite{lu2021learning} generate input functions using a Gaussian process with an RBF kernel, while \cite{boulle2023mathematical} suggest more general covariance structures, such as the Matérn kernel, which provides explicit control over the smoothness of sampled functions. Further discussion on sampling from such distributions is provided in Section \ref{sec:data-gen}.

For theoretical analysis, \cite{lanthaler2022error} considers $ \mathcal{P} $ as the class of distributions whose covariance operators have finite trace norms. This family is related to mean-square continuous processes with covariances that are integral operators of Mercer kernels, as considered by \cite{subedi2024benefits}. Meanwhile, \cite{liu2024deep} adopts a non-parametric perspective, defining $ \mathcal{P} $ as the set of all compactly supported measures on $ \mathcal{V} $.

\section{OPERATOR CLASSES}\label{sec:op-classes}

In this section, we present an overview of select classes frequently used in the operator learning literature. Our primary focus is to describe how mappings between infinite-dimensional spaces are defined and to examine the learning-theoretic aspects of these classes. Thus, this discussion is not meant to serve as an exhaustive review of all classes in operator learning. For a more comprehensive overview, we refer the readers to articles by \cite{boulle2023mathematical} and \cite{kovachki2024operator}.

\subsection{Linear Operators}
For scalar-valued linear regression, the class of linear mappings consists of functions of the form $x \mapsto \langle \beta, x \rangle$.  A straightforward analog of this model can be defined for operator learning. Let $\mathcal{V}$ and $\mathcal{W}$ be Hilbert spaces with orthonormal bases $\{\varphi_j\}_{j \in \mathbb{N}}$ and $\{\psi_j\}_{j \in \mathbb{N}}$, respectively. For a sequence $\beta = \{\beta_j\}_{j \in \mathbb{N}}$, consider the mapping 
\begin{equation}
 v \mapsto \sum_{j=1}^\infty \beta_j \langle v, \varphi_j \rangle  \, \psi_j.
 \label{eq:lin-model-basis}
\end{equation}
One can define a class of such mappings  as $\Fcal = \{v \mapsto \sum_{j=1}^\infty \beta_j \langle v,  \varphi_j \rangle  \, \psi_j \, \, : \, \, \norm{\beta}_{\ell^2(\naturals)} \leq c\}$.  

Since the orthonormal bases are fixed apriori, learning this linear map reduces to estimating the sequence $ \{\beta_j\}_{j \in \mathbb{N}} $. This is effectively a regression task in the frequency domain, a well-studied problem in statistical modeling \citep{harvey1978linear}. Moreover, this estimation framework is closely related to principal component functional regression, which has been studied in works of \cite{hormann2015note} and \cite{reimherr2015functional}. Recently, \cite{de2023convergence} and \cite{subedi2024error} analyzed this model in the context of learning solution operators of PDEs.

Despite its simplicity, this model can be used to approximate the solution operator of the heat equation \ref{eq:heat}. Let $ \{\varphi_j\}_{j=1}^\infty $ be the eigenfunctions of the Laplacian operator $ -\nabla^2 $ on the domain $ \mathcal{X} $, with corresponding eigenvalues $ \{\eta_j\}_{j \geq 1} $. That is, $
-\nabla^2 \varphi_j = \eta_j \, \varphi_j$ for all $j \in \naturals$.
Then, the solution operator of the heat equation has a spectral representation \citep{dodziuk1981eigenvalues} such that any initial condition $u_0$ is mapped to the solution function at time point $t$ as 
\[ u_0 \mapsto \sum_{j=1}^{\infty} e^{-\tau \eta_j  t} \inner{u_0}{\varphi_j} \varphi_j, \]
which can be parametrized as Equation \ref{eq:lin-model-basis}.

A slightly more general linear model can be defined using an integral operator associated with a kernel. Let $k_{\theta}: \mathcal{X} \times \mathcal{X} \to \mathbb{R}^{p \times p}$ be a kernel parameterized by $\theta$. Define the operator $\operatorname{K}_{\theta}$ as follows
\begin{equation}\label{eq:kernel-integral}
    (\operatorname{K}_{\theta}v)(y) = \int_{\mathcal{X}} k_{\theta}(y, x)\, v(x) \, d\nu(x),
\end{equation}
where $\nu$ is a measure on $\mathcal{X}$. This model has been extensively studied in the functional data analysis literature, especially when $p = 1$ and $v \in L^2([0, 1], \mathbb{R})$ \citep[Section 3.2]{wang2016functional}. 

Kernel integral operators are particularly useful for learning linear boundary value problems of the form $\mathcal{L} w = v$, where $w(x) = 0$ for all $x \in \text{boundary}(\mathcal{X})$. Under regularity conditions such as uniform ellipticity, the solution operator for these problems can be represented as an integral operator of the associated Green's kernel \citep{boulle2023elliptic}. For example, when $\Xcal = \reals^d$, the solution operator of the heat equation can also be written as the integral operator of Gaussian kernel \citep[Section 2.3]{evans2022partial}, that is
\[u_t(y) = \int_{\reals^d} \frac{1}{(4\pi \tau t)^{d/2}} \exp\left( -\frac{\norm{y-x}^2}{4 \tau t}\right)\, \, u_0(x)\, dx. \]
However, for general domains $\Xcal$, the heat kernel does not always have a closed-form solution. In fact, deriving an explicit Green’s function is not possible for most linear PDEs, particularly in high dimensions.

To address this, \cite{boulle2022data} proposed learning the Green’s function directly from data by parameterizing it with a neural network. This approach offers two main benefits. First, the problem reduces from estimating an operator on infinite-dimensional spaces to estimating a function defined on a finite-dimensional domain. Second, Green’s functions capture fundamental properties such as conservation laws and symmetries, providing better interpretability compared to the general solution operator. Additionally, when the underlying boundary value problem is non-linear, \cite{gin2021deepgreen} proposed to use an encoder-decoder framework, where the Green function is learned in an encoding space where the PDE is linearized. More recently, \cite{stepaniants2023learning} developed a theoretical framework for learning the Green’s function within a reproducing kernel Hilbert space (RKHS). Using the theory of kernel methods, \citet{stepaniants2023learning} derived rigorous statistical guarantees for the resulting estimator.

\subsection{Neural Operators}
Unlike the heat equation, many PDEs of practical interest have nonlinear solution operators. Neural operators are a class of neural network-based architectures developed to approximate such nonlinear operators. Recall that a multilayer neural network is defined by sequential composition of multiple single layer networks. In standard finite dimensional settings, a single layer network is a map $x \mapsto \sigma(Wx+b)$ for some pointwise non-linearity $\sigma(\cdot).$ Neural operator is a natural generalization of such mapping to function spaces. In particular, for a given input function $v$, a single layer of a neural operator is defined as 
\[ v(y) \mapsto  \sigma \Big((\operatorname{K}_{\theta} v)(y) + b(y) \Big), \quad \text{ where } \quad  (\operatorname{K}_{\theta} v)(y) = \int_{\mathcal{X}} k_{\theta}(y, x)\, v(x) \, d\nu(x). \]
Here, $\operatorname{K}_{\theta}$ is a kernel integral operator of a kernel $k_{\theta}$ parameterized by $\theta$, $b: \mathcal{X} \to \mathbb{R}^p$ is a bias function, and $\sigma: \mathbb{R}^p \to \mathbb{R}^p$ is a pointwise nonlinear activation function, such as ReLU.
We refer interested readers to \cite{kovachki2023neural} for a detailed discussion.

Although the neural operator model in its current form was introduced by \cite{li2020neural, li2020multipole} and further developed in \cite{kovachki2023neural}, the underlying mapping $v(y) \mapsto \sigma \Big((\operatorname{K}_{\theta} v)(y) + b(y)\Big)$ has been studied in the FDA literature as functional single and multi-index models (see \citep[Equation 13]{wang2016functional} and \citep{chen2011single}). However, the operator learning literature has developed novel techniques that enable fast training and efficient evaluation of such nonlinear mappings. 

Since data is typically provided on a grid over a domain of size $N$, a naive implementation of this model based on numerical integration would require a time complexity of $O(N^2)$. 
Since the grid size $N$ is typically exponentially large in $d$, this naive approach becomes a significant bottleneck while training large neural operator models with multiple layers. Thus, recent advances in neural operator methods have focused on overcoming this computational limitation by introducing techniques that reduce the complexity and enable efficient training and evaluation of these models at a large scale. One such technique involves parametrizing the kernel $k_{\theta_t}$ in the Fourier domain, which gives rise to a well-known architecture called the Fourier Neural Operator.

\subsubsection{Fourier Neural Operators (FNO)}\label{sec:fno}

The original formulation of Fourier Neural Operator (FNO) \citep{li2020fourier} assumes a translation-invariant kernel and applies the convolution theorem. Here, we present an equivalent formulation using a Mercer-type decomposition of $ k_{\theta} $. Suppose $ k_{\theta}: \mathcal{X} \times \mathcal{X} \to \mathbb{R}^{p \times p} $ admits the expansion
\[
[k_{\theta}(y, x)]_{ij} = \sum_{m \in \mathbb{Z}^d} \lambda_{ij}(m) e^{2\pi \mathrm{i} m \cdot (y-x)},
\]
where $ \lambda_{ij}: \mathbb{Z}^d \to \mathbb{C} $. Define the matrix-valued function $\Lambda: \mathbb{Z}^d \to \mathbb{C}^{p \times p}$ such that $[\Lambda(m)]_{ij} = \lambda_{ij}(m)$. Using this decomposition and changing the order of sum and integral, the kernel operator $\operatorname{K}_{\theta}$ can be written as
\[
    (\operatorname{K}_{\theta}v)(y) = \sum_{m \in \mathbb{Z}^d} e^{2\pi \mathrm{i} m \cdot y} \, \left( \Lambda(m) \int_{\mathcal{X}} e^{-2\pi \mathrm{i} m \cdot x} \, v(x) \, d\nu(x) \right).
\]
This formulation describes FNOs as first computing the Fourier transform of $ v $, applying a transformation $ \Lambda(m) $ to each mode, and reconstructing the output using an inverse Fourier transform, often written succinctly as $
\operatorname{IFT} \big(\Lambda \cdot \operatorname{FT}(v) \big).
$

There are two key challenges in a practical implementation of this transformation. First, the infinite summation over $ \mathbb{Z}^d $, and second, the approximation of $ \operatorname{FT}(v) $ due to discretization. \cite{li2020fourier} address the first challenge by truncating the sum to $|m|_{\ell^{\infty}} \leq M$, reducing parameters of the model to at most $ M^d $ matrices. The second challenge is addressed by approximating $\operatorname{FT}(v)$ using the discrete Fourier transform (DFT) of $v$, computed over the finite grid of domain points. Suppose the data is available on a uniform grid of size $N$. Then, DFT can be efficiently computed using fast Fourier transform (FFT) algorithms, which have a computational complexity of $O(N \log N)$. 

Note that this parameterization is just a special case of Equation \ref{eq:lin-model-basis}, where the basis is explicitly chosen as the Fourier basis. More generally, kernel expansion can use alternative bases, such as Chebyshev polynomials or wavelets \citep{tripura2023wavelet, gupta2021multiwavelet}, which can be better for non-periodic or non-smooth functions. Variants of FFT also exist for these transformations, though they may require different grid structures such as Chebyshev nodes \citep[Chapter 2]{trefethen2019approximation} or dyadic grids for wavelets \citep[Chapter 7]{mallat1999wavelet}.

\subsubsection{DeepOnet}\label{sec:deeponet}
While neural operators are typically described as architectures composed of mappings of the form $v \mapsto \sigma(\operatorname{K}v + b)$, there are also other ways in which neural network modeling has inspired operator learning architectures. Perhaps the most popular one is the DeepOnet architecture proposed by \cite{lu2021learning}, based on the pioneering work of \cite{chen1995universal}. Given the input function $v$, the DeepOnet architecture is a mapping
\[v(y) \mapsto \sum_{j=1}^{q} b_j\big(\Ecal(v)\big) \, t_j(y).\]
Here, $\Ecal: \Vcal \to \reals^r$ is an encoder that encodes the input function $v$ on $\reals^r$, 
$b_j: \mathbb{R}^r \to \mathbb{R}$ is typically a neural network that processes an encoding of $v$ in $\mathbb{R}^r$, and $t_j: \mathcal{X} \to \mathbb{R}^p$ is another neural network that takes the evaluation point $y$ of the output function in $\Wcal$ as an input. 
Although the learned functions \( t_j \) may not be orthogonal in practice, this architecture can be interpreted as follows: the functions \((t_j)_{j=1}^q\) learn the dominant basis of \(\mathcal{W}\) that captures the most significant variations in the target functions, while the functions \((b_j)_{j=1}^q\) map an input function to the corresponding basis coefficients.

 A common choice for the encoder is $\mathcal{E}(v) = (v(x_1), \ldots, v(x_N))$, where $x_1, \ldots, x_N$ are the grid points at which $v$ is sampled. However, this encoding is not discretization-invariant, meaning the model cannot generalize to evaluate a new input function $v'$ sampled on a different grid. To achieve discretization invariance, $v$ can instead be encoded as $\mathcal{E}(v) = (\langle v, \varphi_1 \rangle, \ldots, \langle v, \varphi_r \rangle)$, where $(\varphi_j)_{j=1}^r$ are orthonormal functions in the space $\mathcal{V}$. In a special case when \( (\varphi_j)_{j=1}^r \) correspond to the first \( r \) principal components of the covariance operator of \( \mu \) and \( (t_j)_{j=1}^q \) represent the first \( q \) principal components of the covariance operator of the pushforward measure \( \operatorname{G}_{\#}(\mu) \), the resulting architecture is referred to as PCA-Net \citep{bhattacharya2021model}.

\subsection{ RKHS and Random Features }

Let  $\mathcal{W}$ be a Hilbert space. Recall that the usual reproducing kernel Hilbert space (RKHS) of scalar-valued functions can be defined using a real-valued kernel.
However, defining a RKHS of operators requires an {\em operator-valued} kernel. To that end, let $\mathcal{B}(\mathcal{W})$ denote the space of bounded linear operators on $\mathcal{W}$. A function $K: \mathcal{V} \times \mathcal{V} \to \mathcal{B}(\mathcal{W})$ is an operator-valued kernel if:  (1) $K$ is Hermitian, i.e., $K(u, v) = K(v, u)^*$ for all $u, v \in \mathcal{V}$ and   (2) $K$ is non-negative, meaning that for any  $\{(v_i, w_i)\}_{i=1}^n$,  we have $
   \sum_{i=1}^n\sum_{j=1}^n \langle w_i, K(v_i, v_j)w_j \rangle_{\mathcal{W}} \geq 0.
$

The class $\mathcal{F}$ is an operator RKHS associated with $K$ if  
\begin{itemize}
    \item[(1)] $K(v, \cdot) w \in \mathcal{F}$ for all $v \in \mathcal{V}$, $w \in \mathcal{W}$. 
    \item[(2)] For any $v \in \mathcal{V}$, $w \in \mathcal{W}$, and $\operatorname{F} \in \mathcal{F}$, we have  $
  \langle \operatorname{F}, K(v, \cdot) w \rangle_{\mathcal{F}} = \langle \operatorname{F}(v), w \rangle_{\mathcal{W}}.
$
\end{itemize}
The second property is the reproducing property, which implies a closed-form solution for penalized least-squares estimation under the RKHS norm. Given training data $\{(v_i, w_i)\}_{i=1}^n$, the solution to such penalized-estimation problem takes the form  
\[
\widehat{\operatorname{F}}_n(\cdot) = \sum_{i=1}^n K(v_i, \cdot) \alpha_i,
\]
for some coefficients $\alpha_1, \ldots, \alpha_n \in \mathcal{W}$. Further details on operator RKHS and the corresponding estimation problem can be found in the work of \cite{micchelli2005learning}. 
The framework of using operator RKHS for function valued regression was developed in a series of works by \cite{kadri2010nonlinear, kadri2011functional, kadri2016operator}. In an interesting work, \cite{batlle2024kernel} showed that these RKHS-based methods for operator learning are competitive with neural network-based methods on standard benchmark datasets. More recently, \cite{mora2025operator} developed a Gaussian process framework for operator learning where they estimate a real-valued measurement of the solution operator, which allows them to use finite-dimensional kernel methods.

Despite the theoretical appeal of the estimator mentioned above, its practical implementation is challenging. The primary challenge arises from the fact that the kernel $K$ is operator-valued.  \cite{kadri2016operator} studied a simplified case where \( K \) takes the form \( K(u, v) = k(u, v) L \) for a scalar-valued kernel \( k \) and a fixed linear operator \( L \). Even under this restrictive assumption, computing the solution requires expensive tensor-based techniques when \( L \) is non-diagonal. For a general kernel \( K \), the computational cost becomes prohibitive for most practical applications.

A more practical alternative to working directly with operator-valued kernels is the random feature model for operators proposed by \cite{nelsen2021random}. The random feature model (RFM) for scalar-valued functions was originally introduced by \cite{rahimi2007random} as a scalable approximation of kernel methods by constructing a randomized low-dimensional approximation of the true RKHS feature map. This idea was later extended to matrix- and operator-valued kernels by \cite{brault2016random} and \cite{minh2016operator}, respectively. However, given the difficulty of defining and implementing non-trivial operator-valued kernels, \cite{nelsen2021random} proposed RFM as a standalone model rather than an approximation technique for kernel methods. Thus, our focus here is only on the resulting trainable random feature model.  For a detailed discussion on operator RFMs and their connection to kernel methods, we refer the readers to \citep{nelsen2021random, nelsen2024operator}.

Defining an RFM requires a feature map $\Phi : \Vcal \times \Theta \to \Wcal$, where $\Theta$ is a set of parameters and a probability measure $ \rho$ on  $\Theta$. Then, a random feature model indexed by $\vartheta$  is 
\[\operatorname{F}_{\text{RFM}}\left(v ; \vartheta \right) := \frac{1}{M}\sum_{j=1}^M \beta_j \Phi(v ; \theta_j) \]
for all $v \in \Vcal$. Here, $\beta_1, \ldots, \beta_M$ are some scalars and $\vartheta := \{\theta_j\}_{j=1}^M$, where $\theta_j$'s are iid draw from the probability measure $\rho.$
The class $\Fcal$ indexed by $\vartheta$ can be defined as all such RFMs such that $\sum_{j=1}^M |\beta_j|^2 \leq c$ for some $c>0$. This model is trained in so-called lazy regime, where the random parameters are fixed and one only trains the scalar parameters $\beta_1, \ldots, \beta_M$. Since this is a linear optimization problem, one can derive the unique minimizer and establish statistical guarantees of the resulting estimator \citep{LN2023randomfeatures}.  

Although RFM was originally conceptualized in \cite{rahimi2007random} from its connection to kernel methods, one can go beyond feature maps of the associated kernels. For example, one can consider a random feature model where $\Phi(\cdot, \theta_j)$'s are random initialization of large powerful models such Fourier Neural Operators or DeepOnets. Such a model can potentially give us the expressivity of neural networks while providing rigorous statistical guarantees.

\section{DATA GENERATION, ESTIMATION, AND EVALUATION}
Beyond the model itself, the success of operator learning depends on the choice of data distribution, estimation strategy, and evaluation framework. We next discuss these key practical aspects of operator learning.

\subsection{Data Generation and Sampling}\label{sec:data-gen}

One of the defining features of operator learning, particularly for applications involving partial differential equations (PDEs), is the flexibility in data acquisition. Unlike traditional machine learning tasks, where data is often collected from fixed experiments or observations, operator learning benefits from the ability to generate labeled data by directly querying a PDE solver. The process begins with sampling input functions $v$ from a carefully chosen distribution $\mu$.

\begin{figure}[h]
    \centering
    \begin{subfigure}[b]{0.32\textwidth}
        \centering
        \includegraphics[width=\textwidth]{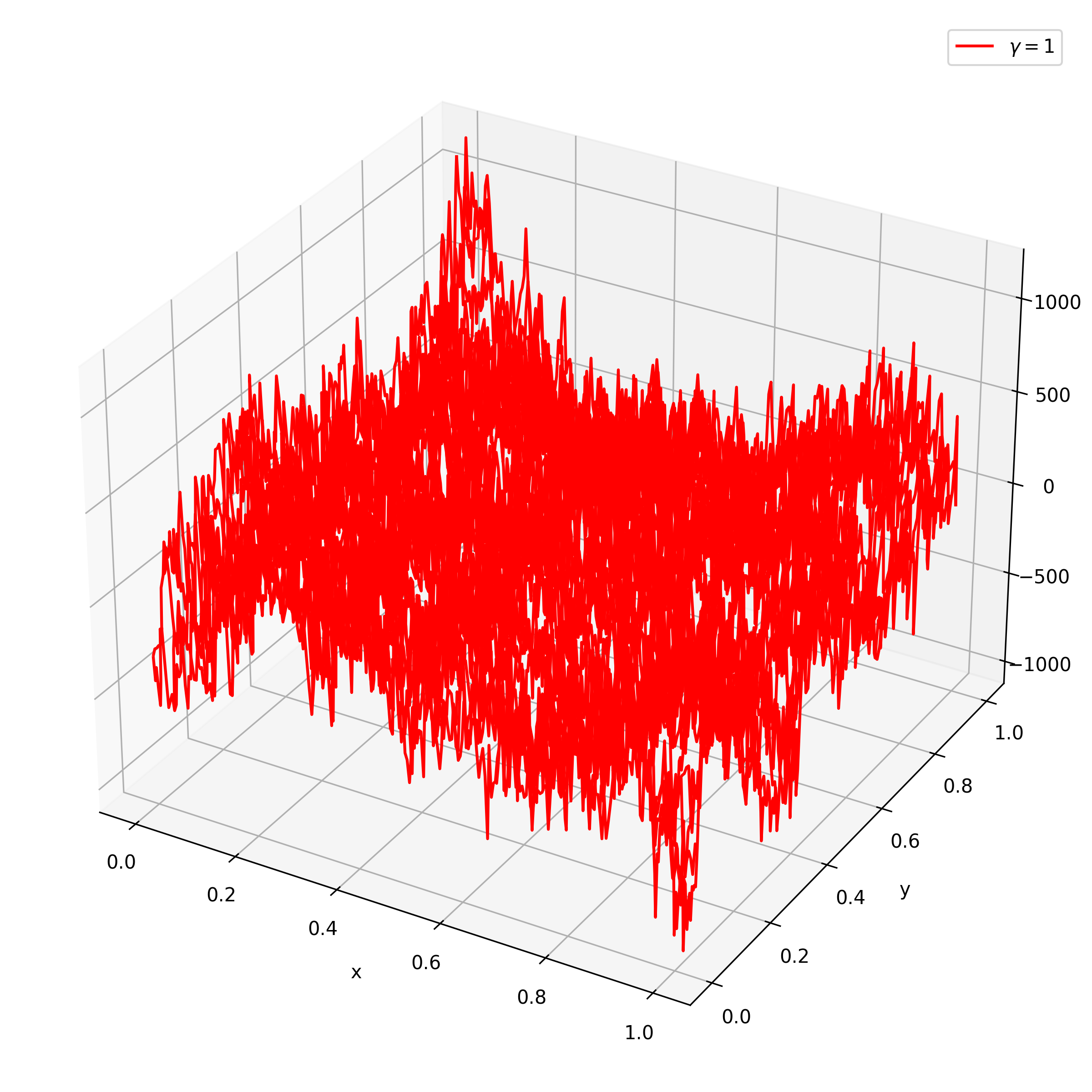}
        \caption{$\gamma=1$}
    \end{subfigure}
    \begin{subfigure}[b]{0.32\textwidth}
        \centering
        \includegraphics[width=\textwidth]{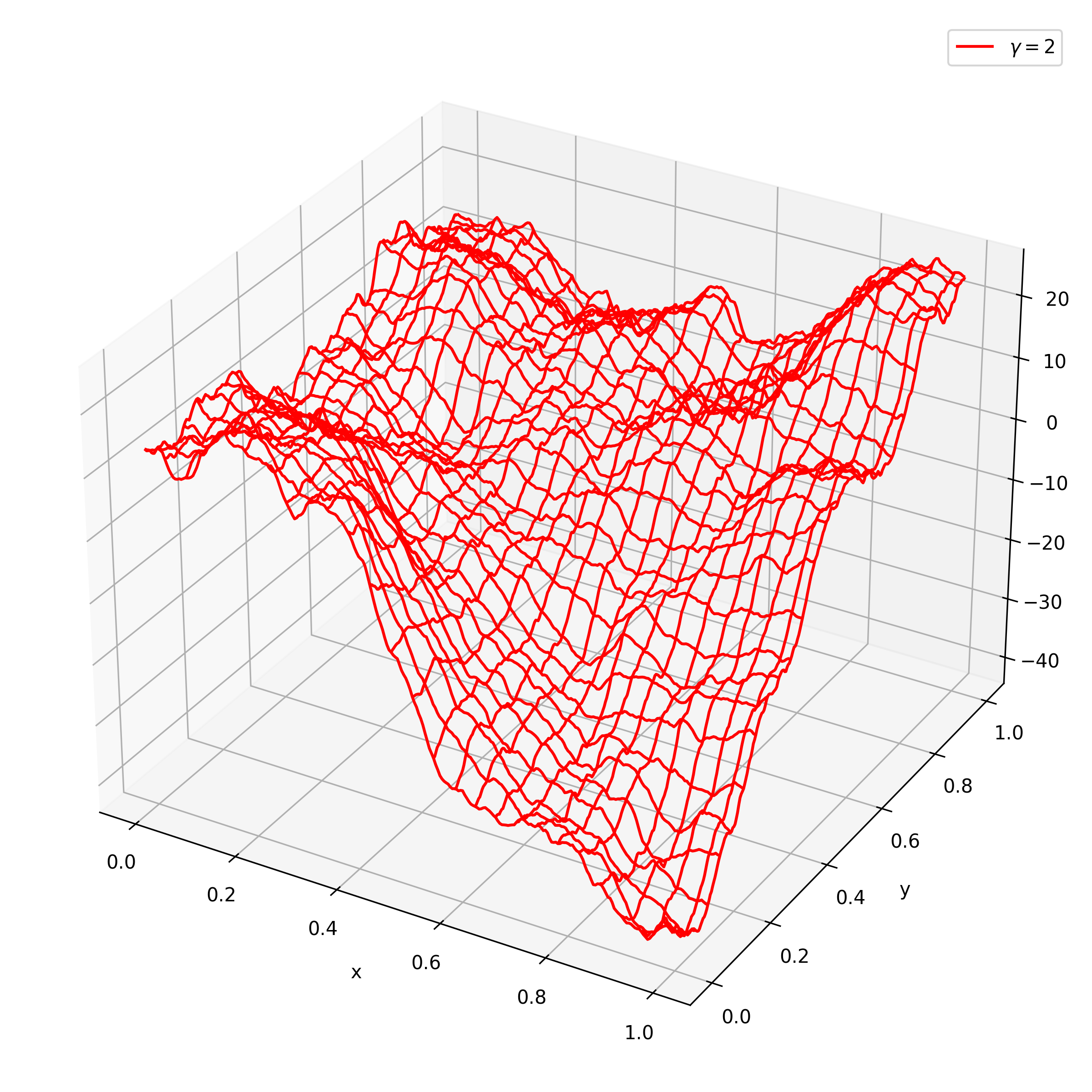}
        \caption{$\gamma=2$}
    \end{subfigure}
    \begin{subfigure}[b]{0.32\textwidth}
        \centering
        \includegraphics[width=\textwidth]{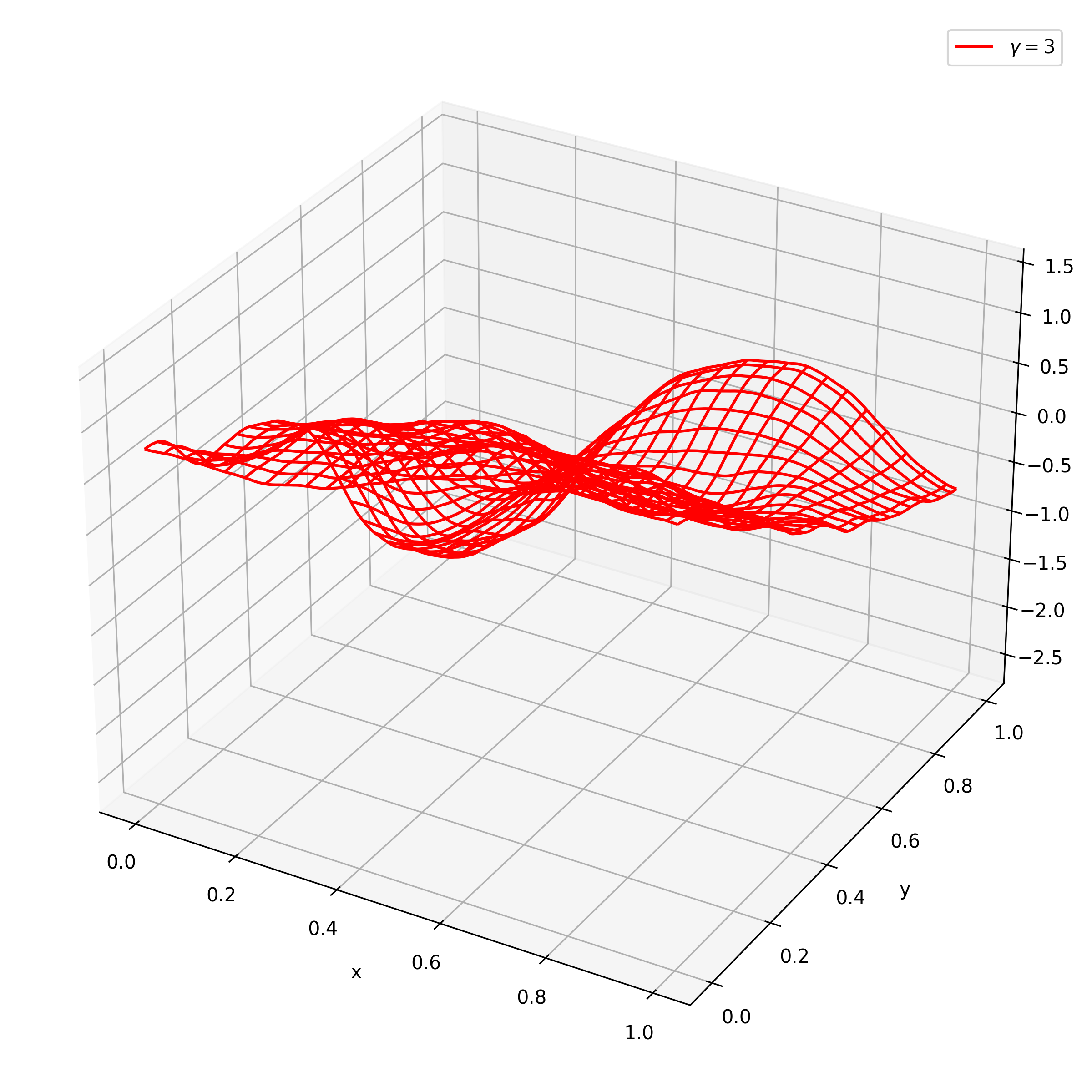}
        \caption{$\gamma=3$}
    \end{subfigure}
    \vspace{0.2cm}
    \caption{Samples from $ \operatorname{GP}\big(0, (-\nabla^2 + \mathbf{I})^{-\gamma} \big) $ for different values of $ \gamma $ on the domain $[0,1)^2$, illustrating how $ \gamma $ determines the smoothness of the generated functions.}
    \label{fig:samples}
\end{figure}

In the applied literature,  $v$ is typically sampled from a mean-zero Gaussian process with a covariance operator of the form $\alpha (-\nabla^2 + \beta \identity)^{-\gamma}$. This distribution, widely used in the applied stochastic PDEs literature \citep{lord2014introduction}, was first proposed for operator learning by \cite{bhattacharya2021model} and also implemented in later works \citep{li2020fourier, kovachki2023neural}.  As Figure \ref{fig:samples} shows, the parameter 
$\gamma$ controls the average smoothness of the generated samples. This allows practitioners to adjust 
$\gamma$ to incorporate prior knowledge about the smoothness of input functions relevant to the specific application.

To sample input functions from such a distribution, one makes use of Karhunen-Loève decomposition. Let $\{\varphi_j\}_{j=1}^\infty$ denote the eigenfunctions of $-\nabla^2$ on $\mathcal{X}$, with corresponding eigenvalues being $(\eta_j)_{j \geq 1}$. By the Spectral Mapping Theorem, $\{\varphi_j\}_{j=1}^\infty$ remain the eigenfunctions of  $(-\nabla^2 + \beta \mathbf{I})^{-\gamma}$, with eigenvalues being $\lambda_j := \alpha(\eta_j + \beta)^{-\gamma}$. If $\sum_{j=1}^\infty \lambda_j < \infty$, then the Karhunen-Loève Theorem \cite[Theorem 7.3.5]{hsing2015theoretical} states that any sample $v \sim \operatorname{GP}(0, \alpha(-\nabla^2 + \beta \mathbf{I})^{-\gamma})$ can be decomposed as
\[
v(x) = \sum_{j=1}^\infty \sqrt{\lambda_j} \, \xi_j \, \varphi_j(x) \quad \quad \forall x \in \mathcal{X},
\]
where $\{\xi_j\}_{j=1}^\infty$ are uncorrelated standard Gaussian random variables on $\reals$. Thus, sampling  $v$ is reduced to sampling a sequence of independent Gaussian random variables $(\xi_j)_{j=1}^\infty$, which is often truncated to  $(\xi_j)_{j=1}^M$ in practice. The resulting truncation yields a sample $
v(x) = \sum_{j=1}^{M} \sqrt{\lambda_j} \, \xi_j \, \varphi_j(x)
$, which is then evaluated on a predefined discrete grid of $\mathcal{X}$.

Sampling from this distribution requires knowledge of eigenpairs \( (\lambda_j, \varphi_j)_{j \geq 1} \), which depend on the domain geometry. In certain cases, these can be computed analytically. For example, on the $1d$ torus \( (\mathbb{T} \simeq [0,1]) \), the eigenfunctions are the Fourier basis, with eigenvalues \( \lambda_j = \alpha(\beta + 4\pi^2 |j|^2)^{-\gamma} \). This also extends to higher-dimensional torus \(\mathbb{T}^d\)  \citep[Section 3.2.1]{subedi2024error}. Similarly, on a \( d \)-dimensional sphere \( S^d \), the eigenfunctions are spherical harmonics, though closed-form expressions may not exist for arbitrary domains.

Moreover, other kernels such as the Matérn or radial basis function (RBF) can also be used. For example, \cite{lu2021learning} generate samples using the RBF kernel, $K(y, x) = \exp(-\|x - y\|^2 / (2\ell^2))$. The eigenfunctions of the RBF kernel are Hermite polynomials with corresponding eigenvalues that decay exponentially fast \cite[Section 4.3.1]{williams2006gaussian}. When the eigenpairs of the kernels are not available in closed form, various numerical methods can be used to approximate them \citep[Section 4.3.2]{williams2006gaussian}. Additionally, it is possible to go beyond Gaussian processes and sample $(\xi_j)_{j \geq 1}$ from other distributions
with heavier tails such as from $t$-distribution or
$\text{Uniform}([-c, c])$ for some $c > 0$.

Once the input function $v$ has been generated, numerical solvers are used to obtain the corresponding solution $w = \operatorname{G}(v)$. There are many numerical methods such as finite difference, spectral, and finite element. To keep this discussion focused on the statistical aspect of operator learning, we will not expand further on these numerical methods. We refer the interested readers to \citet[Section 4.1.2]{boulle2023mathematical} and references therein.

\subsection{Estimation}\label{sec:training}

Given $ n $ i.i.d. samples $ \{(v_i, \operatorname{G}(v_i))\}_{i=1}^n $, the estimator is typically obtained by empirical risk minimization, 
\[
\widehat{\operatorname{F}}_n \in \argmin_{\operatorname{F} \in \Fcal} \frac{1}{n} \sum_{i=1}^{n} \ell(\operatorname{F}(v_i), \operatorname{G}(v_i)).
\]  
As discussed in Section \ref{sec:loss}, common choice of the loss function includes squared $ L^2 $ loss, $
\ell(\operatorname{F}(v_i), \operatorname{G}(v_i)) = \norm{\operatorname{F}(v_i) - \operatorname{G}(v_i)}^2_{L^2} $, or a related relative loss.  Since the data is only available on a discrete grid $ \{x_1, \ldots, x_N\} \subset \Xcal $, the $ L^2 $ norm is approximated empirically as  $
\norm{\operatorname{F}(v_i) - \operatorname{G}(v_i)}_{L^2}^2 \approx \frac{1}{N} \sum_{j=1}^{N} \big| \operatorname{F}(v_i)(x_j) - \operatorname{G}(v_i)(x_j) \big|^2.
$
The optimization problem is then solved using first-order methods such as stochastic gradient descent (SGD) and its variants. When the underlying PDE is known, additional constraints imposed by the PDE such as conservation laws are often incorporated as a regularization term in the optimization problem. This technique, referred to as physics-informed training, is discussed further in Section \ref{sec:pde-specific}.

\subsection{Evaluation \& Out-of-Distribution Generalization}

Suppose $ \widehat{\operatorname{F}}_n $ is the operator estimated from the training data. Its performance is then evaluated on a separate set of held-out test samples $ \{(v_i, \operatorname{G}(v_i))\}_{i=1}^{n_{\text{test}}} $. The test samples may be drawn from the same distribution as the training set or from a different distribution. For example, if the training distribution $ \mu $ is a Gaussian process with covariance $ \alpha(-\nabla^2 + \beta \mathbf{I})^{-\gamma_1} $, the test distribution $ \mu_{\text{test}} $ could have a covariance of the form $\alpha (-\nabla^2 + \beta \mathbf{I})^{-\gamma_2} $ for some $ \gamma_2 < \gamma_1 $. Since the parameter $ \gamma_j $ determines the smoothness of the sampled functions, evaluating on a test set with a smaller $ \gamma_2 $ allows for assessing the model’s ability to generalize to functions with lower smoothness. 

Moreover, as discussed in Section \ref{sec:data-gen}, functions are typically sampled using the Karhunen-Loève decomposition. That is, a function can be expressed as  $
v(x) = \sum_{j=1}^{M} \sqrt{\lambda_j} \, \xi_j \, \varphi_j(x),
$ 
where function samples are generated by drawing values for $ \xi_j $. To assess out-of-distribution generalization, one can modify the distribution from which $ \xi_j $ is sampled. For example, while training data may be generated by drawing $ \xi_j $ from a Gaussian distribution, test samples could be generated using a long-tailed distribution, such as a t-distribution.

\section{ERROR ANALYSIS AND CONVERGENCE RATES}

For a fixed $\mu$, the excess risk $ \mathcal{E}_n(\widehat{\operatorname{F}}_n, \mathcal{F}, \Pcal, \operatorname{G}) $ defined in Equation \ref{eq:excess-risk} consists of two terms: the risk of the estimator $\widehat{\operatorname{F}}_n$ and the risk of the optimal operator in $\mathcal{F}$. The latter term, 
\[
\inf_{\operatorname{F} \in \mathcal{F}} \expect_{(v, w) \sim \mu} \big[ \ell\big(\operatorname{F}(v), \operatorname{G}(v)\big) \big],
\]
is referred to as the \emph{approximation error} of $\Fcal$. This error arises due to model misspecification-- the ground truth $\operatorname{G}$ lying outside the class $\mathcal{F}$. Since this error is irreducible and does not vanish even as $n \to \infty$, it is subtracted from the error of the estimator so that the resulting excess risk can go to zero in the limit. 

In traditional learning problems, the excess risk $\mathcal{E}_n(\widehat{\operatorname{F}}_n, \mathcal{F}, \Pcal, \operatorname{G})$ is typically referred to as the \emph{statistical error} of the estimator $\widehat{\operatorname{F}}_n$. Statistical error arises because the learner must identify the optimal operator in $\mathcal{F}$ for the distribution $\mu$, despite only having access to a finite number of samples from that distribution. However, in operator learning, unlike traditional learning settings, the excess risk contains at least two additional sources of error beyond the statistical error: the \emph{discretization error} and the \emph{truncation error}. The discretization error arises because the learner has access to samples only on a discrete grid of domain points, but is evaluated on the entire domain. On the other hand, the truncation error arises due to various finite-dimensional approximations of the infinite-dimensional object of interest.


\subsection{Approximation Error}
Most theoretical works in operator learning have focused on quantifying the approximation error of various model classes. One of the earliest works was by \cite{chen1995universal}, who introduced a neural network-based architecture called operator networks and established its universality for representing continuous operators on compact subsets. Building on this work, \cite{lu2021learning} proposed DeepONet and proved its universality. While these initial results were qualitative, \cite{lanthaler2022error} provided quantitative error estimates for these architectures.
The universality of Fourier Neural Operators (FNOs) and their associated error bounds were established by \cite{kovachki2021universal}. Similar approximation error estimates were later established for PCA-Net by \cite{lanthaler2023operator}. For a more comprehensive overview of works on approximation error of various operator classes, we refer the readers to \citep[Sections 4 \& 5]{kovachki2024operator}.

\subsection{Truncation Error}
Truncation errors can arise from a variety of sources. For example, as discussed in Section \ref{sec:data-gen}, while samples $v \sim \operatorname{GP}(0, \alpha(-\nabla^2 + \beta \mathbf{I})^{-\gamma})$ has the representation $v = \sum_{j=1}^\infty \sqrt{\lambda_j} \xi_j \varphi_j$, only a truncated sum $\sum_{j=1}^M \sqrt{\lambda_j} \xi_j \varphi_j$ is used in practice. This introduces a truncation error in the data itself. 

Truncation errors also arise during the estimation process. For example, in model \ref{eq:lin-model-basis}, only a finite subset of the parameters \((\beta_j)_{j=1}^\infty\) can be estimated in practice, introducing a truncation error. More generally, in operator learning literature, the model class $\mathcal{F}$ is often assumed to have a low-dimensional latent structure. Specifically, for any $\operatorname{F} \in \mathcal{F}$, there exists a mapping $h: \mathbb{R}^r \to \mathbb{R}^q$ belonging to another class $\mathcal{H}$ such that $\operatorname{F} = \operatorname{D} \circ \,h \circ \operatorname{E}$, where $\operatorname{E}$ and $\operatorname{D}$ are predefined encoder and decoder operators. 
A common choice for the encoder is a finite-dimensional projection, defined as $\operatorname{E}(v) = (\langle v, \varphi_1 \rangle, \ldots, \langle v, \varphi_r \rangle)$, where $(\varphi_j)_{j=1}^\infty$ is an orthonormal basis of the input space $\mathcal{V}$. Similarly, given a vector $\zeta \in \mathbb{R}^q$, the decoder can be defined as $\zeta \mapsto \sum_{j=1}^q \zeta_j \psi_j$, where $\zeta = (\zeta_1, \ldots, \zeta_q)$ and $(\psi_j)_{j=1}^\infty$ is an orthonormal basis of the output space $\mathcal{W}$. Such encoders and decoders naturally introduce a truncation error.
\cite{lanthaler2022error} characterized the encoding and decoding errors for the DeepONet architecture, while a similar analysis for PCA-Net was provided in \cite{lanthaler2023operator}. Furthermore, \cite{liu2024deep} studied truncation error of various encoder-decoder frameworks for the estimation of Lipschitz operators.

Truncation errors are also inherent in FNO estimation (Section \ref{sec:fno}), where the infinite sum is approximated by a finite truncation. \cite{subedi2024benefits} quantify this error for the linear core of FNO. Similarly, PCA-based estimators in functional linear regression \citep{hormann2015note, reimherr2015functional} also incur truncation errors as only a finite number of principal components are used in practice.

\subsection{Discretization Error}\label{sec:discretization-error}
Discretization error fundamentally arises from approximating functional operations using numerical methods on a finite-sized grid. \cite{lu2021learning} quantified the discretization error in sampling from a Gaussian process and reconstructing functions via linear interpolation. For DeepONet, \cite{lanthaler2022error} analyzed the error from encoding $v$ using numerical inner products on a discrete grid. More recently, \cite{lanthaler2024discretization} analyzed discretization error in Fourier Neural Operators (FNOs) during evaluation. For a fixed FNO model \( \operatorname{F} \), they bounded the numerical error of evaluating $\operatorname{F}(v)$ on a grid of size \( N^d \) instead of the exact evaluation.

A more relevant discretization error arises during the learning process itself rather than model evaluation. It occurs because the estimator is trained on a discrete grid of size \(N\) but evaluated at full resolution as \(N \to \infty\). \cite{subedi2024error} quantified this error for learning the linear core of FNOs. However, a more practical analysis would consider training on a grid of size \(N_1\) and evaluating on a different grid of size \(N_2\). Such a general analysis would provide a rigorous foundation for understanding the multiresolution generalization, also known as zero-shot super-resolution, often observed in practice \citep[Section 5.4]{li2020fourier}. This phenomenon refers to the ability of an operator trained on lower-resolution data to exhibit strong performance even when evaluated on higher-resolution grids. Therefore, a key direction for future research is developing a general theory of multiresolution generalization.  

The problem of learning from discretized data shares similarities with partial information settings studied in the bandit literature \citep{lattimore2020bandit}. Thus, techniques from bandit literature could help in understanding learning theoretic consequences of discretization. Additionally, often in practice, training data is available at multiple grid sizes. A natural way to study this could be through the lens of missing data problems, borrowing tools from the statistical literature \citep{little2019statistical}.

\subsection{Statistical Error}

There is a substantial body of work studying statistical error when $\mathcal{F}$ is a class of linear operators. In the FDA literature,  $\mathcal{V}$ is often assumed to be $L^2([0,1])$, the output space $\mathcal{W}$ is typically $\mathbb{R}$, and $\mathcal{F}$ is generally a class of integral transforms. For foundational results in this setting, we refer readers to the seminal works of \cite{hall2007methodology} and \cite{yuan2010reproducing}, as well as the references therein. A non-technical overview of these results is available in review articles such as \cite{wang2016functional} and \cite{morris2015functional}. These results have been extended to settings where the response is function-valued, say $\mathcal{W} = L^2([0,1])$, in works like \cite{yao2005functional}, \cite{crambes2013asymptotics}, \cite{benatia2017functional}, and \cite{imaizumi2018pca}. Further generalizations to arbitrary Hilbert spaces $\mathcal{V}$ and $\mathcal{W}$ have been developed in \citep{hormann2015note, reimherr2015functional, de2023convergence}. All of the works mentioned above generally establish prediction consistency results with associated rates of estimation under the additive noise model $w = \operatorname{F}(v) + \delta$.
In contrast, \cite{tabaghi2019learning} and \cite{subedi2024error} bound excess risk for learning linear operators in potentially fully misspecified (agnostic) settings. It is important to note that this list is far from exhaustive and represents only a subset of works from a mature and extensive body of literature on linear operator learning.

Compared to linear operator learning, the statistical theory for nonlinear operator classes $\mathcal{F}$ is relatively underdeveloped. Early works in this area include works on additive and polynomial models \citep{li2008sharpening, yao2010functional}. \cite{kadri2016operator} extended the theory to the estimation of operators when $\mathcal{F}$ is a reproducing kernel Hilbert space (RKHS) of operators. Additionally, \citet{batlle2024kernel} developed a kernel-based framework for operator learning and established statistical guarantees for scenarios where data is observed as finite-dimensional vectors through a fixed measurement map. \cite{nelsen2021random} generalized the random feature model of \cite{rahimi2007random} to learning operators, and comprehensive statistical analysis for RFM was later provided by \cite{LN2023randomfeatures}. 

Building on the early work of \cite{mhaskar1997neural}, recent work by \cite{kovachki2024data} provides a statistical analysis for non-parametric operator classes such as Lipschitz operators. However, their findings primarily highlight the hardness of the problem, as they show that the minimax error bound decays very slowly, at a rate of  $ \sim (\log n)^{-1}$. A related analysis by \cite{liu2024deep} also reports similarly pessimistic rates for Lipschitz operators estimation. These results suggest that Lipschitz operators might be too big of a class to learn. Thus, an important future direction is to identify a smaller class $ \mathcal{F}$ that captures operators commonly encountered in practice for which efficient statistical estimation is possible.

While most applied works in operator learning rely on neural network-based architectures, their statistical analyses remain limited. \cite{lanthaler2022error} and \cite{gopalani2024towards} established generalization bounds for DeepONets, while \cite{kim2024bounding} and \cite{benitez2024out} derived bounds for Fourier Neural Operators using Rademacher complexity estimates. However, these bounds typically rely on covering number analysis, leading to an exponential dependence on the parameter size or number of layers. Thus, the resulting bounds are too loose to provide any meaningful insights into the empirical success of these models. In future, shifting focus from deep architectures to tighter analyses of simpler models, such as single-layer networks, may yield deeper insights into the statistical properties of these neural network-based operator models.

\subsection{Towards a General Statistical Theory of Operator Learning}

Beyond analyzing specific operator classes, it would be of interest to develop a general statistical theory for operator learning. Historically, statistical learning theory has focused on developing such general frameworks, beginning with the Vapnik-Chervonenkis (VC) theory introduced by \cite{vapnik1971uniform} and further developed through tools from empirical process theory by \cite{dudley1978central}, \cite{talagrand2005generic}, and others. A key challenge in developing a similar theory for operator learning is identifying an appropriate complexity measure for the operator class $ \mathcal{F} $ that effectively quantifies its complexity of learning. Modern statistical learning theory relies primarily on covering numbers to quantify such complexity \citep{van1996weak, van2000empirical}. Recent work by \cite{reinhardt2024statistical} has taken a step in this direction by providing estimation bounds for compact operator classes in terms of their covering number. However, covering numbers may not be the right tool in this setting as many fundamental operator classes are inherently non-compact, making covering number-based bounds vacuous. 

For example, consider a simple class of constant functions 
\[ \mathcal{F} = \{ v \mapsto w \mid w \in \mathcal{W} \text{ and } \|w\|_{\mathcal{W}} \leq 1 \}, \]
which corresponds to the unit ball in the output space $ \mathcal{W} $. Since the unit ball in an infinite-dimensional space is non-compact, the covering number of $ \mathcal{F} $ is unbounded under any reasonable metric. Yet, learning this class is just a reframing of a mean estimation problem. Given $ n $ i.i.d. samples $ \{(v_i, w_i)\}_{i=1}^n $, the empirical mean $ \widehat{\operatorname{F}}_n = \frac{1}{n} \sum_{i=1}^n w_i $ achieves the standard convergence rate of  $ \mathcal{E}_n(\widehat{\operatorname{F}}_n, \mathcal{F}, \mathcal{P}, \operatorname{G}) \lesssim n^{-1/2} $  when $\Wcal$ is a separable Hilbert space and $\ell$ is the canonical norm of $\Wcal$.
This suggests that a new complexity measure other than covering number is needed—one that meaningfully characterizes learnability for operator classes, much like what VC dimension does for binary classification \citep{blumer1989learnability}.  More precisely, given a class $\Fcal$, is there a complexity measure function $\operatorname{C}_{\gamma}(\Fcal)$ such that $\mathcal{E}_n(\widehat{\operatorname{F}}_n, \mathcal{F}, \Pcal, \operatorname{G}) \xrightarrow{n \to \infty} 0$ if and only if $\operatorname{C}_{\gamma}(\Fcal) <\infty$ for every $\gamma>0$?  In other words, the measure $\operatorname{C}_{\gamma}(\Fcal)$ tells us when the risk-consistent estimation of $\Fcal$ is possible. A promising direction could be to develop generalizations of the fat-shattering dimension, which has been used to characterize learnability in scalar-valued regression \citep{bartlett1996fat, alon1997scale}. The theory of scalar-valued regression, developed in terms of the fat-shattering dimension, suggests that the appropriate complexity measure \( \operatorname{C}_{\gamma}(\Fcal) \)  also provides a tight bound on the rate at which \( \mathcal{E}_n(\widehat{\operatorname{F}}_n, \mathcal{F}, \Pcal, \operatorname{G}) \) converges to zero.

\section{PDE-SPECIFIC OPERATOR LEARNING}\label{sec:pde-specific}
Existing works in operator learning typically use neural networks as black-box estimators for solution operators, which makes them broadly applicable across various PDEs. However, these methods often neglect structural properties inherent to specific PDE classes, leading to suboptimal data efficiency. To address this, a few works have attempted to integrate mathematical or physical constraints into the learning process, often referred to as \emph{physics-informed learning}. The central idea is to encode known physical constraints (e.g., boundary conditions, conservation laws) directly into the training objective as a regularization term. Specifically, a physics-informed approach minimizes a regularized loss function of the form
\[
\widehat{\operatorname{F}} \in \argmin_{\operatorname{F} \in \mathcal{F}} \left( \frac{\lambda_1}{n} \sum_{i=1}^n \ell(\operatorname{F}(v_i), w_i) + \lambda_2 R(S_n, \operatorname{F}) \right)
\]
for some prespecified weights  $\lambda_1,\lambda_2>0 $. Here, $S_n$ is the training sample set, and $R: S_n \times \mathcal{F} \to [0, \infty]$ is a regularization functional encoding PDE priors such as boundary conditions \citep{li2024physics} or variational formulations \citep{goswami2023physics}).

In our running example of the heat equation \ref{eq:heat}, suppose $ u $ is integrable and vanishes on the boundary. It is well known that the total heat energy is conserved, meaning $
\int_{\mathcal{X}} u_t(x)\, dx = \int_{\mathcal{X}} u_0(x)\, dx, \quad \forall t > 0.
$
A natural way to enforce this constraint in operator learning is through a regularization functional
\[
R(S_n, \operatorname{F}) := \frac{1}{n} \sum_{i=1}^n \left| \int_{\mathcal{X}} \big(\operatorname{F}(v_i)(x) - v_i(x) \big)\, dx \right|.
\]
This penalizes deviations from the heat conservation law to ensure that the learned operator respects the underlying physical principle.

The use of physics-based constraints in learning models for solving PDEs was formally introduced within the framework of Physics-Informed Neural Networks (PINNs) by \cite{raissi2019physics}.  By restricting the search space from $ \mathcal{F} $ to operators $ \operatorname{F} $ that satisfy physical laws, the effective model complexity is reduced. This generally leads to improved sample efficiency.

While physics-informed learning is an important first step, a more effective approach could be to incorporate these constraints directly into the architecture design, leading to PDE-specific models. For example, consider the time-dependent Schrödinger equation,  
\[
\imaginary \hbar \frac{d \psi_t}{dt} = \operatorname{H} \psi_t.
\]
The development of surrogate models for Schrödinger’s equation has been an active area of research since the 1970s, with many research communities still dedicated to this pursuit. Thus, an off-the-shelf neural network is unlikely to make a meaningful progress. However, a carefully designed operator learning approach tailored specifically to the Schr\"odinger equation itself could be a valuable addition to the existing toolbox of approximate methods.

The structure of this PDE itself can be used to inform architecture design. For example, when the Hamiltonian $ \operatorname{H} $ is time-independent, the solution operator takes the form $ \operatorname{G} = \exp(-\imaginary \frac{t}{\hbar} \operatorname{H}) $, which is unitary, that is $ \operatorname{G}^{\dagger} \operatorname{G} = \identity $. Thus, a more data-efficient learning approach may involve parametrizing $ \mathcal{F} $ such that every $ \operatorname{F} \in \mathcal{F} $ is unitary. Extending existing parameterizations of unitary matrices, such as one by \cite{jarlskog2005recursive}, could provide a useful starting point.

Finally, we end by noting that a similar PDE-informed design philosophy underlies the Green function learning framework proposed by \citet{boulle2022data} and \citet{gin2021deepgreen}, where the architecture is derived from the observation that solution operators of certain boundary value problems can be expressed as integral operators with Green’s functions. Extending this type of principled approach of architectural design to general nonlinear PDEs is an important future direction.

\section{FUTURE DIRECTIONS}

While operator learning holds great potential, scaling it for real-world applications presents key challenges. We now outline some important research directions to address these issues.

\subsection{Active Data Collection}\label{sec:active}
It is unclear if the iid-based statistical model is the right framework to study operator learning for PDEs. This is because the learner can generate any training data by querying the numerical solver, and thus has no reason to be limited to iid samples from some source distribution. In fact, as generating training data requires computationally expensive numerical solvers, the learner \emph{should} ideally generate data adaptively to ensure that the computational cost of training is justified by saving during evaluation. The model where the learner can adaptively select the data is referred to as active learning model in statistical learning theory. We will propose an active learning model and argue why this is the right model to study operator learning for surrogate modeling of PDE. 

Given a sample size budget of $n$, the learner can pick \emph{any} functions $v_1, v_2, \ldots, v_n \in \Vcal$ and get the label $\operatorname{G}(v_i)$ for each $i \in [n]$. Note that this framework where the learner can request labels for any input is generally unrealistic in many problems. For example, for human data, it may not be feasible to request a label for an individual with an arbitrary feature vector, as such a representative human may not exist in reality. However, this is perfectly realistic in operator learning because the PDE solver can provide a solution for any input function in the appropriate function space $\Vcal$.

With this actively collected training data $ (v_i, \operatorname{G}(v_i))_{i=1}^n$, the learner has to produce an estimate $\widehat{\operatorname{F}}_n$ such that
\[     \mathcal{E}_n^{\text{active}}(\widehat{\operatorname{F}}_n, \mathcal{F}, \Pcal, \operatorname{G}) := \sup_{\mu \in \Pcal} \left( \expect_{v_1, \ldots, v_n} \left[\expect_{v \sim \mu} \big[ \ell\big(\widehat{\operatorname{F}}_n(v), \operatorname{G}(v)\big) \big] \right] - \inf_{\operatorname{F} \in \mathcal{F}} \expect_{v \sim \mu} \big[ \ell\big(\operatorname{F}(v), \operatorname{G}(v)\big) \big] \right)\]
vanishes to $0$ as $n \to \infty$. The key distinction between $\mathcal{E}_n^{\text{active}}(\widehat{\operatorname{F}}_n, \mathcal{F}, \mathcal{P}, \operatorname{G})$ and the excess risk $\mathcal{E}_n(\widehat{\operatorname{F}}_n, \mathcal{F}, \mathcal{P}, \operatorname{G})$ defined in Equation \ref{eq:excess-risk} lies in the definition of $\widehat{\operatorname{F}}_n$. In Equation \ref{eq:excess-risk}, $\widehat{\operatorname{F}}_n$ is constructed using labeled samples where the inputs $v_i$ are drawn independently from $\mu$. In contrast, for $\mathcal{E}_n^{\text{active}}(\widehat{\operatorname{F}}_n, \mathcal{F}, \mathcal{P}, \operatorname{G})$, the learner has the flexibility to select any $v_1, \ldots, v_n \in \mathcal{V}$. The expectation over $v_1, \ldots, v_n$ accounts for any randomness introduced by the learner in the data generation process.

This seemingly minor distinction can have significant implications for the guarantees that can be established in this setting. For example, when $\Fcal$ is a class of linear operators, \cite{subedi2024benefits} showed that fast convergence rates of $n^{-\beta}$ for $\beta \gg 1$ can be obtained for many natural families $\Pcal$. This is in stark contrast to the i.i.d. setting, where rates faster than $n^{-1}$ are not possible. Therefore, this result highlights the substantial advantage of active data collection over i.i.d. sampling, at least for linear operator learning. Recent empirical work by \cite{musekamp2024active} also highlights the benefits of active data collection in operator learning.  \cite{li2024multi} also showed empirically that an active learning strategy that jointly selects input functions and their resolution improves the data efficiency of FNOs. 

The statistical benefits of active learning over passive sampling in traditional learning problems are well-established in the learning theory literature \citep{hanneke2013statistical, settles1994active}. Given this potential benefit, developing active data collection strategies and rigorously establishing their statistical benefits is an important direction for future research in operator learning,  as emphasized by \citet{azizzadenesheli2024} in his ICML 2024 tutorial.

\subsection{Uncertainty Quantification}
For operator learning to be deployed in real-world applications, particularly in safety-critical domains, a rigorous framework of uncertainty quantification is essential. Several works have already explored this direction, mostly using Bayesian techniques. For example, \cite{zou2025uncertainty} uses Hamiltonian Monte Carlo to sample from the posterior distribution, using the mean for point prediction and the variance as an uncertainty estimate. Similarly, \cite{magnani2022approximate, akhare2023diffhybrid, guo2024ib} propose approximate Bayesian methods for uncertainty quantification, while \cite{ma2024calibrated} proposes conformal prediction to provide distribution-free coverage guarantees. 

Beyond ensuring model reliability, uncertainty quantification can actually be used to improve the operator learning pipeline in two ways. First, it allows uncertainty-guided active data collection, which can potentially improve sample efficiency by prioritizing high-uncertainty inputs for labeling. A standard Bayesian active learning strategy involves estimating the empirical posterior covariance to define an acquisition objective
\[
U_n(v) = \frac{1}{m} \sum_{i=1}^m \norm{\operatorname{F}_i(v) - \frac{1}{m} \sum_{j=1}^m \operatorname{F}_j(v)}_{\Wcal}^2,
\]
where $ \operatorname{F}_1, \dots, \operatorname{F}_m $ are posterior samples from $ n $ training data points. Then, the next input to label is selected as $ v_{n+1} = \argmax_{v \in \Vcal} U_n(v) $, often from a predefined candidate set $S \subset \Vcal$ of finite size for computational efficiency. This is particularly useful in settings where acquiring labeled data is costly, such as physical experiments, where retraining the model is much cheaper than running a new experiment. A similar approach to uncertainty-guided active learning using ensemble predictions was proposed by \cite{musekamp2024active}.

Second, uncertainty quantification allows robust decision-making in downstream tasks where operator surrogates are integrated into optimization pipelines. For example, in design optimization \citep{rao2019engineering}, the surrogate prediction $ \widehat{w} $ is used to minimize a downstream objective function $ J(\xi, w) $ over a feasible parameter space $ \Xi $, subject to design constraints $ C(\xi) \leq 0 $. A robust optimization can account for uncertainty by solving a min-max problem,
\[
\argmin_{\xi \in \Xi} \max_{w \in B(\widehat{w})} J(\xi, w) \quad \text{subject to} \quad C(\xi) \leq 0,
\]
where $ B(\widehat{w}) $ is an uncertainty set derived from Bayesian posterior samples or conformal prediction. 

Thus, a proper uncertainty framework not only improves model reliability but also allows the development of more sample-efficient and robust systems. As such, developing reliable and scalable uncertainty quantification methods is a crucial future direction in operator learning.

\subsection{Local Averaging and Ensemble Methods}
Given a labeled dataset \( \{(v_i, w_i)\}_{i=1}^n \), local averaging methods such as nearest neighbors or Nadaraya-Watson kernel smoothing define an estimator \( \widehat{\operatorname{F}}_n \) as
\[ \widehat{\operatorname{F}}_n(v) = \sum_{i=1}^n \widehat{\alpha}_i(v)\, w_i, \]
where \( \widehat{\alpha}_i: \mathcal{V} \to [0,1] \) are weights satisfying \( \sum_{i=1}^n \widehat{\alpha}_i(v) = 1 \) for every \( v \in \mathcal{V} \). This method outputs a weighted average of the training labels, with weights determined by the new input. Prior works have established the consistency of $k$-nearest neighbors for functional regression with both scalar response \citep{laloe2008k} and functional response \citep{lian2011convergence} as well as the consistency of Nadaraya-Watson estimators \citep{aspirot2009asymptotic}. However, their empirical performance in surrogate modeling of PDEs has not yet been studied. These methods may provide competitive alternatives to parametric models, particularly in data-scarce settings. Additionally, their ability to update efficiently makes them well-suited for settings where the data is acquired sequentially.

Ensemble methods, like local averaging, compute the weighted averages of multiple predictions. Formally, they define an estimator
\[ \widehat{\operatorname{F}}_n(v) = \sum_{j=1}^m \widehat{\beta}_j \, \widehat{\Psi}_j(v), \quad \forall v \in \mathcal{V}, \]
where \( \widehat{\Psi}_j: \mathcal{V} \to \mathcal{W} \) are base predictors trained on all or part of the dataset \( \{(v_i, w_i)\}_{i=1}^n \), and \( \widehat{\beta}_j \in [0, \infty) \) are learned weights. Common ensemble techniques include bagging, boosting, and random forests.

Ensemble methods, especially boosting, have been shown to outperform neural networks in various applications, particularly when data is scarce, heavy-tailed, skewed, or highly variable \citep{mcelfresh2023neural}. Given that operator learning for PDEs is a data-scarce settings and PDEs on complex geometries introduce data irregularities, ensemble methods may provide strong alternatives to neural operators. Functional regression literature already provide a foundation on these methods, with works on boosting \citep{ferraty2009additive, tutz2010feature} and implementations like FDboost \citep{brockhaus2020boosting}, as well as works on bagging for functional covariates \citep{secchi2013bagging, kim2022bootstrap} and functional random forests \citep{moller2016random, rahman2019functional}. While some methods handle functional targets (see \citep[Section 7]{brockhaus2020boosting}), most are designed for scalar-valued outputs. Extending these techniques to Banach space-valued targets is a useful starting point. Moreover, since existing functional ensemble methods are typically limited to small datasets, scaling them using computational techniques from operator learning \citep{kovachki2023neural} is another natural direction. Finally, rather than replacing neural operators, ensemble methods could also complement them as an ensemble of smaller neural operator models may outperform a single large one.

\subsection{Frictionless Reproducibility}
A key future direction for advancing operator learning is fostering what \cite{donoho2024data} describes as frictionless reproducibility-- a research environment where scientific findings, computational experiments, and learning models can be easily replicated, verified, and built upon with minimal effort. 

A foundational step in this direction is the development of standard benchmark datasets. \cite{lu2022comprehensive} took an important step by releasing 16 benchmark datasets. Additionally, the \texttt{NeuralOperator} library \citep{kovachki2023neural, kossaifi2024library} has a few datasets for benchmarking. However, most existing datasets are on toy problems, which, while valuable for proof-of-concept validation, do not fully capture the complexity of real-world applications. Thus, there is a clear need to expand existing benchmarks to include large-scale datasets for applications such as climate modeling, materials science, and molecular dynamics, where operator learning is expected to have a major impact. Additionally, a single large-scale dataset could significantly accelerate progress: for example, consider the central role ImageNet \citep{deng2009imagenet} played in transforming computer vision.

Beyond datasets, another critical component of frictionless reproducibility is open-source implementations of operator learning models. Some progress has already been made in this direction with the development of libraries such as \texttt{DeepXDE} \citep{lu2021deepxde} and \texttt{NeuralOperator} \citep{kossaifi2024library, kovachki2023neural}. Additionally, many independent works have publicly released their models and datasets. However, these models and datasets are often scattered across different repositories, making it difficult for researchers to locate and systematically compare them. To address this, a centralized platform for hosting operator learning datasets and models would be of huge benefit to the community. Existing platforms such as Hugging Face could be used for sharing datasets, pre-trained models, and benchmarking results in a way that allows seamless collaboration and reproducibility.

\subsection{Scaling Sample and Model Size}  
Recent works on scaling laws \citep{kaplan2020scaling, zhai2022scaling} suggest that model performance continues to improve as sample size, model size, and computational resources increase.  However, most existing works in operator learning are in a data-scarce setting, with models that are much smaller than those used in computer vision and language modeling.  Thus, a key future direction is to explore how we can train a large-scale operator learning model on massive datasets. There are two key challenges in scaling operator learning methods. First, obtaining labeled data is expensive since obtaining each sample requires running a computationally expensive solver or conducting costly experiments. Second, widely used architectures such as DeepONet and FNOs do not scale as efficiently as transformer-based models, limiting the size of models that can be trained with available resources.  

One approach to mitigating data limitations is training a foundation model on diverse PDE datasets and fine-tuning it for specific tasks, as explored by \cite{subramanian2024towards}. On the model scaling front, transformer-based architectures have been proposed for operator learning \citep{cao2021choose, hao2023gnot, litransformer}, but their quadratic computational cost with grid size limits scalability. Developing more efficient tokenization strategies for transformers to enable large-scale operator learning without excessive computational overhead remains an important future direction.

\section{CONCLUSION}
In this article, we reviewed recent developments in operator learning, emphasizing its statistical foundations and connections to functional data analysis (FDA).  That said, the overall goal of the FDA differs slightly from that of operator learning. In FDA, the focus
is on statistical inference, typically using RKHS-based frameworks. As a result, FDA methods do not always scale to
large datasets. In contrast, operator learning primarily aims at prediction, with an emphasis on creating computationally efficient methods that can be used to
train large models and handle large datasets. However, bridging these fields could be mutually beneficial. FDA's theoretical tools can be used for the analysis of operator learning methods, while methodological advances in operator learning can improve the scalability of FDA techniques.     

Applied research has been the primary driver of advancements in operator learning, and this trend will likely continue. However, unlike other machine learning fields, there is a unique opportunity for theory to catch up more rapidly due to the structured nature of operator learning problems. Here, the ground truth operator is well-defined, and the learner typically has partial prior knowledge along with a strong oracle access to the operator via a numerical solver.
This is in contrast to language or vision modeling, where a ground truth function may not even exist or the learner may only have limited oracle access through available samples. As discussed in Section \ref{sec:pde-specific}, this structure in operator learning allows for designing PDE-specific architectures and methods, which may also introduce further mathematical regularity that enables easier theoretical analysis. Moreover, a key reason for the gap between empirical performance and theoretical lower bounds in modern machine learning is the reliance on the iid assumption, which often fails to reflect practical scenarios where practitioners actively select the mosy informative samples. While the iid model historically became standard as a reasonable tradeoff between practical relevance and analytical tractability, operator learning allows for greater flexibility in data acquisition. This opens the door to alternative learning models that better align with real-world data collection while being theoretically as tractable as the iid framework. Studying such alternative learning models can potentially bridge the gap between theory and practice.

\section*{ACKNOWLEDGMENTS}
We acknowledge the support of NSF via grant DMS-2413089.


\bibliographystyle{plainnat}

\bibliography{references}

\end{document}